\pdfoutput=1

\documentclass[11pt]{article}
\usepackage[dvipsnames]{xcolor}

\usepackage{ACL2023}

\usepackage{times}
\usepackage{pgfplots}
\usepackage{latexsym}
\usepackage{algorithm}
\usepackage{pgfplots}
\usepackage{algpseudocode}
\usepgfplotslibrary{external}
\usepackage{adjustbox}
\usepackage{booktabs}
\usepackage{comment}
\usepackage{enumitem}
\usepackage{tikz}
\setlist{nosep}
\usepackage[T1]{fontenc}

\usepackage[utf8]{inputenc}

\usepackage{microtype}

\usepackage{inconsolata}

\usepackage{multirow}
\usepackage{fontawesome}

\usepackage{arydshln}

\newcommand{\fmicro}{F1\textsubscript{micro} }
\newcommand{\fmacro}{F1\textsubscript{macro} }
\newcommand{\Fmicro}{F1\textsubscript{micro}}
\newcommand{\Fmacro}{F1\textsubscript{macro}}

\title{SheffieldVeraAI at SemEval-2023 Task 3: Mono and multilingual approaches for news genre, topic and persuasion technique classification}

\author{Ben Wu$^\# $, Olesya Razuvayevskaya$^\#$, Freddy Heppell$^\# $, João A. Leite$^\#$, \\ {\bf Carolina Scarton, Kalina Bontcheva}, and {\bf Xingyi Song} \\
Department of Computer Science, The University of Sheffield, Sheffield, UK\\
\texttt{\{bpwu1, o.razuvayevskaya, frheppell1, jaleite1\}@sheffield.ac.uk}}

\begin{document}

\maketitle
\def\thefootnote{\#}\footnotetext{Equal contribution, listed randomly.}\def\thefootnote{\arabic{footnote}}
\begin{abstract}
This paper describes our approach for SemEval-2023 Task 3: Detecting the category, the framing, and the persuasion techniques in online news in a multilingual setup. For Subtask 1 (News Genre), we propose an ensemble of fully trained and adapter mBERT models which was ranked
joint-first for German, and had the highest mean rank of multi-language teams. For Subtask 2 (Framing), we achieved first place in 3 languages, and the best average rank across all the languages, by using two separate ensembles: a monolingual RoBERTa-MUPPET\textsubscript{LARGE} and an ensemble of XLM-RoBERTa\textsubscript{LARGE} with adapters and task adaptive pretraining. For Subtask 3 (Persuasion Techniques), we trained a monolingual RoBERTa-Base model for English and a multilingual mBERT model for the remaining languages, which achieved top 10 for all languages, including 2nd for English. For each subtask, we compared monolingual and multilingual approaches, and considered class imbalance techniques. \footnote{Our code is available at \url{https://github.com/GateNLP/semeval2023-multilingual-news-detection}}
\end{abstract}

\section{Introduction}
With the rise of opinion-manipulating news and misinformation surrounding COVID-19, elections and wars, the task of propaganda and hyperpartisan detection has received much attention over the last five years. Since 2019, various SemEval tasks have addressed detecting hyperpartisan \cite{kiesel-etal-2019-semeval}, sarcasm \cite{abu-farha-etal-2022-semeval}, and persuasion techniques in textual and multimodal data \cite{da-san-martino-etal-2020-semeval, dimitrov-etal-2021-semeval}. This task \citep{semeval2023task3} can be seen as an extension of the latter two tasks, suggesting an expanded ontology of persuasion techniques and addressing other related aspects of persuasion, such as satire, opinionated news, and framing detection.

The three subtasks presented in this shared task are the detection of: 1) genre: opinion, objective reporting or satire; 2) framing techniques: 14 multilabel frames; 3) persuasion techniques: 23 multilabel techniques, which can be grouped into 6 high-level classes.



The data consists of labelled training and development sets in English, French, German, Italian, Polish and Russian, and unlabelled test sets in the same languages plus three zero-shot languages: Spanish, Greek and Georgian.

The main contributions of this paper are twofold: 1) evaluation of the viability of monolingual versus multilingual models for each of the subtasks; and 2) presentation of the models which ranked first in four subtask-language pairs, top three in 16 subtask-language pairs, and were within the top 10 for all.







Our approaches for the three subtasks differ, therefore we present each subtask separately in sections 3-5 respectively. An overview of the techniques used in each subtask is shown in Table \ref{tab:overview}.




\section{Background}

\newcommand{\yes}{{\color{ForestGreen}{\faCheck}}}
\newcommand{\no}{{\color{Gray}{-}}}

\begin{table*}[t]
    \centering
    \vspace{-4mm}
    \scalebox{0.8}{
        \begin{tabular}{l l l l l l l l l}
            \toprule
            Tasks & Text Clean & External Data & Oversampling & Class Weights & Adapters & TAPT & Unseen languages & Ensemble  \\
            \midrule
            Subtask1 & \yes $^\star$ & \yes & \yes $^\star$ & \no & \yes $^\dagger$  & \no & Zero-shot & \yes \\ 
            Subtask2 & \yes & \no & \no & \yes & \yes  & \yes & $\Rightarrow$ EN & \yes \\
            Subtask3 & \no & \no & \no & \yes & \no & \no & $\Rightarrow$ EN & \no \\
            \bottomrule
        \end{tabular}
    }
    \vspace{-2mm}
    \caption{An overview of approaches used. TAPT: Task-adaptive Pre-training, $^\star$ not used in adapter model for submission. $^\dagger$ as part of ensemble. }
    \label{tab:overview}
    \vspace{-4mm}
\end{table*}


Fine-grained propaganda technique classification was first introduced by \citet{da-san-martino-etal-2019-fine}, who suggested a multi-granularity network, where the lower and higher granularity tasks refer to the fragment and sentence-level classification respectively. 
Other state-of-the-art approaches to this task used an ensemble of RoBERTa models with class weighting, where some models perform a semi-supervised task of span detection \cite{jurkiewicz-etal-2020-applicaai} and an ensemble of 5 different transformer models \cite{tian-etal-2021-mind}, namely BERT \cite{devlin-etal-2019-bert}, RoBERTa \cite{liu2019roberta}, XLNet \cite{yang2019xlnet}, DeBERTa \cite{he2020deberta} and ALBERT \cite{lan2019albert}.

Framing detection specifically has been addressed primarily for political news, with the models exploring unsupervised probabilistic topic models combined with autoregressive distributed-lag models \cite{tsur-etal-2015-frame}, finetuning BERT \cite{liu-etal-2019-detecting} and multilingual BERT (mBERT) \cite{akyurek-etal-2020-multi}. The latter system is the closest to our task since it explores the multilabel multilingual setting and the effect of translating texts for use in monolingual models. However, it uses article headlines instead of the full texts as the classification data. 
The authors found that English BERT\textsubscript{BASE} uncased trained on translated data and tested on the data in the target language often outperforms the multilingual model. We perform similar comparison experiments for all three subtasks of this shared task.

\citet{wang-banko-2021-practical} performed a series of experiments comparing monolingual and multilingual approaches for hate speech detection and sentiment analysis and found that different task-language combinations favour either monolingual and multilingual settings. The authors also concluded that data augmentation in the form of translation and task-adaptive pretraining (TAPT) \cite{gururangan-etal-2020-dont} helps to further improve the results.

Another important task addressed in fake news detection is satire detection, with the methods ranging from convolutional neural networks (CNNs) \cite{guibon2019multilingual} to adversarial training \cite{mchardy-etal-2019-adversarial} and BERT-based architectures with long-short-term memory (LSTM) \cite{pandey2022bert, liu2021research} and CNN \cite{kaliyar2021fakebert} layers on top.

\paragraph{Bottleneck Adapters} Adapters \citep{houlsby2019parameter,  bapna-firat-2019-simple} represent a family of techniques aimed at improving parameter efficiency in finetuning by freezing a pretrained model and inserting low-dimension adapter modules within each layer. \citeauthor{houlsby2019parameter} found that, despite training only 3.6\% of the parameters compared to a full model, performance only decreased by 0.4\%, while \citeauthor{bapna-firat-2019-simple} found that adapters produced comparable, or in some cases better, results. Particularly relevant for our task is \citet{he-etal-2021-effectiveness}'s finding that adapter-based tuning of LLMs is particularly effective for low-resource and cross-lingual tasks. For our system, we used two different configurations of the bottleneck adapter modules: 1) the original \citeauthor{houlsby2019parameter} bottleneck configuration, which places adapter modules after the multi-head attention block and feed-forward block of each transformer layer; 2) the \citeauthor{pfeiffer-etal-2020-mad} configuration, which places adapter modules only after the feed-forward block of each layer.  

\citet{chalkidis-etal-2021-multieurlex} also found that for XLM-R on the MultiEURLEX dataset, training bottleneck adapters outperforms traditional full finetuning and improves zero-shot cross lingual capability. Similarly to this task, the MultiEURLEX dataset was used for multilingual multilabel classification, though it is significantly larger than the data in this task (covering 23 languages and classifying hierarchically from 21 to 567 labels).








\section{System Description for Subtask 1}
\subsection{System Overview}\label{sec:st1_system}
The system consisted of an ensemble of four models, comprising 1) three mBERT models each finetuned using the organiser training set and 2/3 of the development set; 2) one frozen mBERT model with a finetuned Houlsby adapter\footnote{Using a data split described in section~\ref{sec:t1_setup}}. The ensemble predictions were decided by majority vote, with rare tie cases handled by selecting the model with the best validation performance.


\paragraph{Full Finetuning} mBERT\textsubscript{BASE} \citep{devlin-etal-2019-bert} was finetuned on a shuffled combination of all languages. Different to previous approaches \cite{wu-dredze-2020-languages, adebara-etal-2020-translating}, we chose the epoch with the best validation performance per-language, instead of overall, since the best overall epoch is not necessarily the best for a given language\footnote{See Appendix \ref{ap:4}}.
By using three identically-configured models in the ensemble, the data sacrificed for model selection can be rotated between them, so overall no data is truly unseen.

\paragraph{Adapter Model} A Houlsby bottleneck adapter was applied to a pretrained mBERT\textsubscript{BASE} model, with a reduction factor of 8 (i.e. $d=96$), using the AdapterHub \citep{pfeiffer-etal-2020-adapterhub} framework. The mBERT model parameters were frozen, so only the adapter and classification head parameters were trained.

\paragraph{Data Preprocessing}
Since task data is obtained from webpages, it often contains unwanted content, such as hyperlinks, account handles, dates and author biographies. We applied the preprocessing described in Appendix \ref{ap:1} to remove this content.

\paragraph{Long Article Truncation}
The organiser annotation instructions indicate that even human annotators find it difficult to distinguish \emph{opinion} from \emph{reporting} and \emph{satire}. This is due to subtle differences in how opinionated direct speech could be balanced or reported on. The instructions also mention that single opinionated sentences, which would trigger the \emph{opinion} genre, often appear at the end of an article. Given the limit on input length for BERT models, for the articles that were longer than 512 tokens, we sequentially selected sentences from the beginning and the end of the article, preserving the original sentence order, until the length of 512 tokens was reached.

\paragraph{External Satire} Due to the lack of satire data, our training set was supplemented with 203 English-language satire articles from \citet{golbeck18-fakenews}.

\paragraph{Data Oversampling} The training data is severely imbalanced, with less than 6\% of articles annotated as \emph{satire} and 18\% of articles annotated as \emph{reporting}. Although the external satire data improved the balance for English, the performance of the satire class in other languages still remained inadequate. To address this, we performed oversampling for both \emph{satire} and \emph{opinion} classes by repeating random oversampling without replacement on the original data for a given language and class until the classes were balanced. For English \emph{satire} class, we applied the same approach but oversampled the external satire data mentioned earlier rather than the original training set. We also compared the effectiveness of the oversampling approach with the class weighting approach in our experiment, and the results showed a slight advantage for the oversampling approach on average.


	
	
	
	
	

\subsection{Experimental Setup}\label{sec:t1_setup}
For the final submission, mBERT transformer models were finetuned on the organiser training set and a part of the development set for 30 epochs with the learning rate of 1e-5, AdamW optimiser ($\varepsilon$=1e-8) and ReLU activation function. The organiser development set was split into three parts, stratified by label and language. We then finetuned three models, using $\frac{1}{3}$ of the development set as a test and merging $\frac{2}{3}$ of the development set with the training data and shuffling the dataset. The held-out part of the development set was used to identify the language-specific best checkpoints for each model. We utilised the checkpoint with the best overall \fmacro on the held-out set to make predictions on the surprise languages. As described in section~\ref{sec:st1_system}, all articles were preprocessed and the training data was oversampled for \emph{satire} and \emph{reporting} classes of each language.


Adapter models were trained on the combined organiser and development set (resplit 80\% train, 10\% validation, 10\% held-out test, stratified by label), for 20 epochs with the learning rate of 1e-4, AdamW optimiser ($\varepsilon$=1e-8) and Tanh activation function in the classifier, and selected the checkpoint with the highest overall validation F1\textsubscript{macro} score. The above preprocessing and oversampling were not used, and articles were truncated at 512 tokens.

After submission, we conducted additional experiments using the organiser training and development sets for consistency. For monolingual models, all articles in the training set and the external satire were translated with Google Translate into the language of each monolingual model in question. Due to the character length limitation, particularly long articles were translated sentence-by-sentence. 

\subsubsection{Results and Reflections}\label{sec:t1_results}
The final submission results of the ensemble are listed in Table~\ref{tab:st1_final_results}.


\begin{table}[]
    \centering
    \scalebox{0.8}{
        \begin{tabular}{l r r  l r r }
           \toprule
           Language  & \fmacro & Place &  Language  & \fmacro & Place\\ \midrule
           English  & 61.282 & 3 & Italian & 72.040 & 3 \\
           French & 68.157 & 5 & Polish & 76.455 & 3 \\
           German & 81.951 & $^\star$1 & Russian & 72.871 & 2 \\ \midrule
           Spanish & 44.293 & 4 & Greek & 68.681	& 6 \\
           Georgian & 96.268 & 2 & & & \\
           \bottomrule
        \end{tabular}
    }
    \vspace{-2mm}
    \caption{Subtask 1 final leaderboard results. $^\star$ joint.}
    \vspace{-4mm}
    \label{tab:st1_final_results}
\end{table}

The final ensemble results achieved a higher \fmacro score than in the supplementary multilingual results in all languages, except Polish. In English, the ensemble achieved \fmacro score 25\% higher than the single mBERT transformer or adapter model. It should be noted that the final models were trained on both training and development data. However, since the development sets are only $\approx\frac{1}{3}$ the size of the training sets, the difference in the amount of the training data was not dramatic.

 In the absence of gold standard labels for the test set, it is difficult to analyse why the model achieved a high score in Georgian, despite being zero-shot. However, our ensemble predictions suggest that there is likely to be no satire articles in the Georgian test set, which was consistently the most difficult class to detect.

 Table~\ref{tab:st1-mono-vs-multi} shows the differences between monolingual and multilingual versions of adapter bottleneck and transformer models, evaluated against the organiser development set. The multilingual transformer models always perform better than monolingual ones, while for 4 out of 6 languages, adapter bottleneck models benefit from the monolingual setup. This may be due to using a fixed reduction factor across all languages. Interestingly, the mBERT model demonstrates the best average result in English for both transformer and bottleneck adapter models. For Italian, XLM-R yields the best results for both transformer and adapter bottleneck models. It is also notable that the results for English are by far the worst across all the models, possibly because the models are overly focused on capturing semantic meaning and are not as effective in genre classification. 
 
 Even though transformer XLM-R demonstrated significantly better results than transformer mBERT for Italian and German, these differences were only marginal in our main setting where the validation set was smaller, while the marginally better results for Russian were not observed at all. Given the above observations and the fact that XLM-R yielded higher \fmacro fluctuations, sometimes reaching 10\%, we opted for the mBERT model as our main submission.

\begin{table*}
    \centering
    \scalebox{0.8}{
        \begin{tabular}{l  r r r r r r}
            \toprule
             \multirow{2}{*}{Language} & \multicolumn{3}{c}{Transformer} & \multicolumn{3}{c}{Adapter} \\ \cmidrule(lr){2-4} \cmidrule(lr){5-7}
             & Monolingual & mBERT & XLM-R & Monolingual & mBERT & XLM-R  \\
            \midrule
            English & 30.0 ±  5.6 & $^*$\textbf{36.2 ±  2.5} & 36.1 ±  2.1 & 20.5 ±  3.3 & $^*$21.4 ±  6.0 & 20.0 ±  3.1\\
            French  & 51.2 ±  3.3 & 62.5 ±  4.6 & $^*$65.5 ±  4.3 & $^*$\textbf{68.3 ±  0.6} & 64.2 ±  0.8 & 61.3 ±  2.7\\
            German  & 59.9 ±  4.1 & 59.9 ±  5.0 & $^*$\textbf{66.9 ±  1.0} & $^*$65.7 ±  3.6 & 57.8 ±  2.8 & 62.0 ±  3.2 \\
            Italian & 56.7 ±  6.5 & 55.1 ±  4.3 & $^*$\textbf{72.6 ±  6.4} & 51.9 ±  4.7 & 47.8 ±  2.3 & $^*$60.3 ±  3.1\\
            Polish  & 71.7 ±  6.6 & $^*$\textbf{81.9 ±  3.7} & 79.4 ±  1.1 & $^*$77.6 ±  2.9 & 72.9 ±  5.1 & 76.7 ±  2.2\\
            Russian & 52.8 ±  8.8 & 52.9 ±  1.6 & $^*$54.7 ±  9.8 & $^*$\textbf{56.9 ±  9.5} & 48.3 ±  2.1 & 48.0 ± 0.8\\
            \bottomrule
        \end{tabular}
    }
    \vspace{-2mm}
    \caption{Mean \fmacro $\pm$ 1 std (over 3 runs) on subtask 1 organiser development set for multilingual and monolingual models for transformer and adapter-only architecture. $^*$ denotes the best per model per language and \textbf{bold} denotes the overall best per language.}
    \vspace{-4mm}
    \label{tab:st1-mono-vs-multi}
\end{table*}

\subsection{Post-competition Findings}
Since, in our final submission, all languages were evaluated without translation (including the three surprise languages), a natural question we wanted to explore after the competition was whether translating texts into a different language for evaluation (the `translate-test' approach) would have yielded better results.

We selected the checkpoints that, during training,  achieved the highest validation \fmacro for each individual language. These `language-optimal' checkpoints were then used to evaluate translations of the other test sets. For example, using the `French'-optimal checkpoint, we translated all tests sets into French and made predictions. 


Surprisingly, we found that the translate-test English $\rightarrow$ Russian and Italian $\rightarrow$ French each improved the \fmacro performance by 1\% on English and Italian respectively, while French $\rightarrow$ Russian improved by over 6\%. 

Two out of the three surprise languages, Spanish and Greek, also benefited from being translated into other languages and tested using the corresponding best checkpoints. The Spanish $\rightarrow$ English setting showed particularly striking increase in \fmacro, from 68.7 to 81.7, which is also 21\% higher than the score of the winning team for Spanish and is 20\% above the other post-competition results. Except for German, translating the Greek test set into the other 5 main languages and testing using the corresponding checkpoints also provided significant improvements in the range of 5\%-13\%, which is over 1\% above the result of the winning system and is the current leaderboard-best.

\section{System Description for Subtask 2}
\subsection{System Overview}
Two systems were used for submission, depending on the language. For English and the three surprise languages, we used \textbf{a monolingual English ensemble} of 3 RoBERTa-MUPPET\textsubscript{LARGE} models. For the remaining languages (French, German, Italian, Russian, Polish), we used \textbf{a multilingual ensemble} of 3 XLM-R\textsubscript{LARGE} models (with adapters and task-adaptive pre-training (TAPT)).

An overview of the two models is shown in Table~\ref{tab:st2_summary}.  
A key difference between these two systems is that the monolingual MUPPET models were trained using traditional finetuning of all parameters, whereas the XLM-R models 1) underwent task-adaptive pre-training; and 2) were finetuned using Pfeiffer bottleneck adapters. 

For both systems, we trained our models jointly on articles in all languages (using English translations for our monolingual model). This meant that we produced a single monolingual or multilingual system that was able to make predictions for all languages. 
We chose this approach of joint training across all languages in order to maximise the number of examples seen for each class, since the dataset for Subtask 2 is quite small, particularly when split by language. Our early experiments showed that this approach was superior to training models on the individual articles in each language. 


\begin{table} 
    \center
    \scalebox{0.8}{
        \begin{tabular}{p{4cm} p{4cm}} 
    
        \toprule 
        MUPPET ensemble & XLM-R ensemble \\ 
        \midrule 
        MUPPET\textsubscript{LARGE} & XLM-R\textsubscript{LARGE} \\
    
        
        Monolingual (English) & Multilingual \\
        Trained on all articles (in translation) & Trained on all articles (original) \\
    
    
        No TAPT & TAPT \\
    
        Full finetuning & Adapter finetuning \\

        Ensemble size 3 &  Ensemble size 3 \\ 
    
        \midrule
        Submitted for & \\

        EN, EL, KA, ES & All other languages \\
        \bottomrule
        \end{tabular}
    }
    \vspace{-2mm}
    \caption{Summary of the monolingual vs multilingual systems submitted for subtask 2}
    \label{tab:st2_summary}
    \vspace{-4mm}
\end{table}

\subsubsection{The English Monolingual System}

\paragraph{Full Finetuning}
Our monolingual system used a finetuned RoBERTa-MUPPET\textsubscript{LARGE} \citep{aghajanyan-etal-2021-muppet} ensemble. RoBERTa-MUPPET improves on its baseline RoBERTa counterpart by adding an additional `pre-finetuning' stage of multi-task learning. 
We opted not to use adapters, because our cross validation experiments showed this worsened performance (see Appendix Table \ref{tab:st2_full_crossvalidation_results}).

\paragraph{English Translations}
Because RoBERTa-MUPPET is a monolingual model, we translated all articles into English for training, and used them for finetuning alongside the original English articles. We performed inference on non-English languages by translating the articles into English: a `translate-test' approach.


\subsubsection{Multilingual System}
We used XLM-RoBERTa\textsubscript{LARGE} \citep{conneau-etal-2020-unsupervised} for our multilingual model. Our system uses two techniques to improve performance: TAPT and adapter layers. 

\paragraph{Task-adaptive Pre-training}
We performed task-adaptive pre-training on the entire XLM-R model, following the approach suggested by \citet{gururangan-etal-2020-dont}. Masked-language modelling was performed, using all available articles (including the organisers' development and test sets). 
We trained for 60 epochs with a learning rate of 1e-4 and batch size of 128 (for full hyperparameters, see Appendix Table~\ref{tab:st2_tapt_hyperparams}). 

TAPT could alternatively be performed by freezing the base model and training the adapters with an MLM objective. Despite being a faster approach, this has been found to sometimes decrease performance \cite{kim-etal-2021-revisiting}.

\paragraph{Adapters}
Our multilingual system used the Pfeiffer bottleneck adapter configuration, with a reduction factor of 8, which for XLM-R\textsubscript{LARGE} corresponds to a bottleneck hidden size of 128.

Although using adapters did result in slightly improved performance, we found that their main advantage came from their low parameter number, which allowed for faster training and more experimentation. 

\subsection{Ensemble}
Predictions made by our ensembles were decided by a majority vote. Each ensemble consisted of 3 individual models (MUPPET models for monolingual; adapter-finetuned XLM-R + TAPT for multilingual). Within each ensemble, two models were trained with class-weighting, and one-without.



\paragraph{Class Weighting} 
Class weighting helped to account for class imbalance by balancing the impact of under- and over-represented classes. When calculating the loss, the logit for each class was multiplied by a class weight that was inversely proportional to the frequency of that class in the dataset.


Overall \fmicro scores were similar for models with or without class weights. Class-weighting did help to improve performance on less frequent frames (such as \textit{Cultural Identity} and \textit{Public Opinion}), but at the expense of more frequent classes (such as \textit{Political}). Additionally, class weights were problematic in the joint language setting, causing varying performance across languages while maintaining similar overall \fmicro. (A comparison for XLM-R is provided in the Appendix Table~\ref{tab:st2_full_crossvalidation_results}.) For this reason, we chose to use a mix of class-weighted and non-class-weighted models for our ensembles in order to reduce the variance of our final systems.









\subsection{Experimental setup}
\subsubsection{Data Preprocessing}
For both monolingual and multilingual models, we cleaned and preprocessed the article text using a set of steps described in appendix~\ref{ap:1} and truncated it to the first 512 tokens. For monolingual English models, we used Google Translate to produce English translations. 

\subsubsection{Data Split}
For subtask 2, we merged the organiser training and development sets, and used 3-fold cross-validation (stratified by language) to identify the best model configurations. We then produced final models by training on the entire training and organizer development set (rather than a 2/3 fold). 

Although this meant we did not have a validation set to judge the final models that went into our ensemble, it enabled training on all available data, which was important due to the small size of the dataset.  

\subsubsection{Hyperparameters}
 We finetuned our monolingual models for a fixed 20 epochs using a learning rate of 3e-5 (warm-up ratio 0.1; linear decay), a batch size of 8, and the AdamW optimiser. 

We used the same hyperparameters for adapter finetuning, except we raised the learning rate to 1e-4. In general, adapters require a higher learning rate than traditional finetuning, and this is reflected by the findings of \citet{chalkidis-etal-2021-multieurlex} for a similar task. 


\begin{table*}
\centering
    \scalebox{0.8}{
    \begin{tabular}{l c c c c c c c}
    \toprule
    \textbf{Monolingual English} & EN & DE & FR & IT & PO & RU & Overall \fmicro \\

    \midrule

    RoBERTa-Large & 68.4 ± 2.0 & 63.5 ± 2.0 & 57.9 ± 2.9 & 60.9 ± 0.2 & 65.8 ± 3.4 & 54.5 ± 2.7 & 63.6 ± 0.1\\
    MUPPET-Large & \textbf{70.4 ± 2.0} & 62.1 ± 3.7 & \textbf{59.0 ± 0.9} & 58.3 ± 1.5 & 65.7 ± 0.9 & 52.9 ± 1.7 & 63.5 ± 0.7\\
    \midrule
    \textbf{Multilingual} & & & & & & & \\
    \midrule
    XLM-R & 68.3 ± 1.4 & 64.4 ± 1.4 & 58.5 ± 0.7 & 60.6 ± 0.5 & 66.5 ± 3.3 & 54.9 ± 2.0 & 64.0 ± 1.2\\
    XLM-R + TAPT + Adapters & 68.2 ± 0.9 & \textbf{65.0 ± 1.8} & 58.5 ± 2.8 & \textbf{61.0 ± 0.6} & \textbf{66.7 ± 3.0} & \textbf{55.7 ± 3.1} & \textbf{64.2 ± 0.3}\\
    \bottomrule
    \end{tabular}
    }

\vspace{-2mm}
\caption{Mean \fmicro scores of class-weighted models (over 3-fold cross-validation). For complete version with ablations and other configurations, see Table \ref{tab:st2_full_crossvalidation_results}}
\label{tab:st2_crossvalidation_results}
\vspace{-4mm}

\end{table*}

\subsubsection{Cross-validation Findings and Language Selection}
Table \ref{tab:st2_crossvalidation_results} displays a condensed summary of our cross-validation results (for full version, see Appendix Table~\ref{tab:st2_full_crossvalidation_results}).\footnote{This table shows the average performance of individual models trained during cross-validation, and not the performance of any ensembles.}
\footnote{The Overall \fmicro column refers to the \fmicro of the entire validation fold and not the mean \fmicro score across languages.} 


For monolingual models, MUPPET\textsubscript{LARGE} achieves an \fmicro of 70.4 on English, outperforming the RoBERTa baseline by 2 points. Similarly, our XLM-R + TAPT + Adapters demonstrates small but consistent improvements over the multilingual XLM-R baseline across most languages. 

When comparing across monolingual and multilingual models, we see that for English, XLM-R models are unable to compete with the performance of monolingual MUPPET. (They are, however, able to match the performance of their monolingual counterpart RoBERTa). In contrast to this, the multilingual models generally demonstrated better performance on non-English languages. This is reflected by the overall \fmicro scores: MUPPET's 63.5 vs XLMR+TAPT+Adapter's 64.2. Based on these results, we decided to use MUPPET for English and XLM-R for other languages. 

For the unseen languages, we decided to use the monolingual `translate-test' approach based on additional holdout experiments that indicated better MUPPET performance. Although this decision enabled us to achieve 1st place on the evaluation test set for Greek and Georgian, our post-competition findings (section \ref{section:st2_post_comp_findings}) discovered that submitting our multilingual model may have achieved even better results.  





\subsection{Results and Reflections}

\begin{table}
    \centering
    \scalebox{0.8}{
    \begin{tabular}{l r r}
       \toprule
       Language  & Test \fmicro & Place \\ \midrule
       Monolingual MUPPET  & & \\
       \midrule
       English  & 57.895 & 1 \\
       \hdashline
       Spanish$^*$ & 50.829 & 3 \\
       Greek$^*$ & 54.630	& 1 \\
       Georgian$^*$ & 65.421 & 1 \\

       \midrule
       Multilingual XLM-R & & \\
       (+ TAPT + Adpt) & & \\
       \midrule
       French & 53.425 & 3 \\
       German & 65.251 & 3 \\
       Italian & 57.079 & 7 \\
       Polish & 64.516 & 2 \\
       Russian & 44.144 & 2 \\
       \bottomrule
    \end{tabular}
    }
    \vspace{-2mm}
    \caption{Subtask 2 final leaderboard results for monolingual and multilingual systems. $^*$Translated to English}
    \label{tab:st2_final_results}
    \vspace{-4mm}

\end{table}



    


The scores and positions of our model are shown in Table \ref{tab:st2_final_results}.


The strong performance of our monolingual model, which achieved first place in 3 out of 4 languages submitted, suggests that the 'translate-test' approach is competitive for performing multilingual classification, especially for a zero-shot cross-lingual scenario. 

Although our multilingual model also performed well, the competition results suggest that even better performance may be achieved by applying our monolingual approach to other languages i.e. for each language, training a native monolingual model with translations. 

Article truncation poses a limitation on our findings, since news frames can potentially be located in parts of the article that are unseen by the model. One of important future extension of these experiments would be to apply long-document processing techniques to this task.

\subsection{Post-competition Findings} 
\label{section:st2_post_comp_findings}
\subsubsection*{ Is a `translate-test' monolingual model really better than a multilingual model?}

After the competition ended, we wanted to compare the performance of our two ensembles across all test set languages, as each system had only been submitted for a subset of languages. 

Surprisingly, in contrast to our cross-validation experiments, we found that for English, our multilingual system outperformed our monolingual submission: 58.475 (multi) vs 57.895 (mono). The multilingual system also performed better on two of the surprise languages: Greek (58.0 vs 54.63) and Spanish (52.023 vs 50.829). In contrast to the findings of \citet{xenouleas2022realistic}, who found translation-based approaches ``vastly outperform cross-lingual finetuning with adapters'', this suggests that the two approaches are competitive with each other. It is difficult, however, to draw firm conclusions from this finding due to the small size of the unseen test set as well as the impact of TAPT. For more details, see Appendix \ref{ap:2}.

\section{System Description for Subtask 3}
\subsection{System Overview} \label{sec:st3-system-overview}

In subtask 3, our focus was not only to maximize the overall \fmicro, but also to ensure a balanced model performance across all classes\footnote{We further discuss this in appendix \ref{ap:s3-full-leaderboard}}. 
Due to the highly imbalanced nature of the 23 classes in subtask 3, achieving a balanced model performance across all classes is challenging.
To address this issue, similarly to subtask 2, we explored cross-lingual training and implemented class weighting. We also explored an oversampling technique, but it did not provide any additional benefit compared to class weighting and increased the training time. 

In contrast to the organiser baseline, which discards paragraphs without a label, we assigned them a class vector of zeros and included them in the training set. This method lead to a significant improvement in performance across all languages\footnote{A comparison of these approaches are in appendix \ref{ap:classless}.}.


We evaluated the performance of the models on unseen languages by training multilingual models while holding out each language. However, the performance significantly decreased for every language when held out. Therefore, we translated the articles of the three unseen languages into English for the final predictions.

Our final models were a \textbf{RoBERTa\textsubscript{BASE}} for the submission of English and translated unseen languages, and an \textbf{mBERT\textsubscript{BASE}} for the remaining languages.

\subsection{Experimental Setup}


 \textbf{RoBERTa\textsubscript{BASE}} was finetuned for 20 epochs only on English data\footnote{Unlike subtask 2, we finetuned the model only with the original English dataset without adding other translated languages.}, with a batch size of 32, truncated to 256 tokens, and AdamW optimizer with a learning rate of 0.00005, 20\% of the training steps as linear warm-up and weight decay of 0.1. A classification threshold of 0.4 was used.

\textbf{mBERT\textsubscript{BASE}} was finetuned on all languages combined and shuffled, with the same hyperparameters as the RoBERTa\textsubscript{BASE} model, except a batch size of 16. 


We conducted an experiment to explore cross-lingual training
by comparing monolingual vs. multilingual models for each language with fixed hyperparameters, across three runs.
The monolingual models were finetuned and validated on only one language, using either RoBERTa-Base for English or mBERT for the other languages, while the multilingual version was an mBERT finetuned on all the languages and validated on each language separately.

For the final submissions, we merged the training and development sets and finetuned the models without validation data. The model with the lowest training loss across all epochs was selected as the best model. We used the random seed that produced the best \fmicro in the previous development set experiments while training the final model.

\subsubsection{Results} \label{sec:st3-results}
Table \ref{tab:st3_final_results} presents the final leaderboard results for subtask 3. Although the main metric for this subtask is \Fmicro, we highlight that our placings for subtask 3 improve considerably when measuring through \Fmacro, placing top 5 in all languages except the three surprise languages and first place for Italian and French, as shown in Table \ref{tab:st3-final-results-micromacro} in the Appendix.

Results for the monolingual and multilingual experiments on the official development set are displayed on Table \ref{tab:st3-dev-f1micro}, where we report the \fmicro for each language. Scores are reported over three runs on different random seed initialisations. Notice that English is the only language in which the monolingual model outperforms the multilingual version.

\begin{table}[]
\begin{center}
\scalebox{0.8}{
\begin{tabular}{lcc}
\toprule
\multirow{2}{*}{Language} & \multicolumn{2}{c}{\fmicro $\pm$ 1 std}     \\ \cmidrule(lr){2-3} 
 & Monolingual         & Multilingual         \\ \midrule
English & \textbf{36.2 ± 0.3} & 31.8 ± 0.6           \\
French  & 40.5 ± 0.4          & \textbf{43.4 ± 0.4}  \\
German  & 36.9 ± 0.5          & \textbf{40.9 ± 0.7}  \\
Italian & 43.4 ± 1.3          & \textbf{47.5 ± 0.3}  \\
Polish  & 28.9 ± 0.8          & \textbf{30.2 ± 0.8}  \\
Russian & 31.5  ± 1.3         & \textbf{37.5  ± 1.7} \\ \bottomrule
\end{tabular}
}
\vspace{-2mm}
\caption{Monolingual vs. multilingual \fmicro scores on the development set for each language on subtask 3. Best \fmicro per language are marked as \textbf{bold}.}
\label{tab:st3-dev-f1micro}
\vspace{-4mm}
\end{center}
\end{table}

Table \ref{tab:st3-zeroshot} shows the \fmicro for the zero-shot experiments, in which a multilingual model was finetuned on all languages except the one being evaluated. By comparison with Table \ref{tab:st3-dev-f1micro}, we see that zero-shot drastically hinders the performance on all languages. For this reason, we decided to translate the test sets for the three surprise languages into English and perform inference using the English monolingual model. We were unable to experiment with translating into different languages other than English due to time constraints.

\begin{table}[]
\begin{center}
\scalebox{0.8}{
\begin{tabular}{@{}ll@{}}
\toprule
\multirow{2}{*}{Language} & \fmicro $\pm$ 1 std \\ \cmidrule(lr){2-2}
    & Zero-Shot      \\ \midrule
English & 21.6 ± 0.4  \\
French  & 35.8 ± 0.7  \\
German  & 28.0 ± 0.6  \\
Italian & 34.3  ± 0.4 \\
Polish  & 21.7 ± 0.6  \\
Russian & 18.6 ± 1.1  \\ \bottomrule
\end{tabular}
}
\vspace{-2mm}
\caption{Subtask 3 \fmicro for the zero-shot experiments.}
\vspace{-4mm}
    \label{tab:st3-zeroshot}
\end{center}
\end{table}


\begin{table}[]
    \centering
    \scalebox{0.8}{
    \begin{tabular}{l r r}
       \toprule
       Language  & Test \fmicro & Place \\ \midrule
       Monolingual RoBERTa\textsubscript{BASE} & & \\
       \midrule
       English  & 36.802 & 2 \\
       \hdashline
       Spanish$^*$ & 27.497 & 9 \\
       Greek$^*$ & 17.426	& 7 \\
       Georgian$^*$ & 24.911 & 10 \\

       \midrule
       Multilingual mBERT & & \\
       \midrule
       French & 41.436 & 4 \\
       German & 44.726 & 6 \\
       Italian & 52.494 & 3 \\
       Polish & 34.700 & 7 \\
       Russian & 31.841 & 5 \\
       \bottomrule
    \end{tabular}
    }
    \vspace{-2mm}
    \caption{Subtask 3 final leaderboard results for monolingual and multilingual systems. $^*$Translated to English.}
    \vspace{-4mm}
    \label{tab:st3_final_results}
\end{table}

\section{Conclusion}
We presented three systems aimed at solving three subtasks within SemEval-2023 Task 3. Our systems applied a variety of state-of-the-art techniques including adapters and TAPT, and consistently achieved a high rank across all available languages, including zero-shot low-resource languages. We additionally presented an analysis of the viability of monolingual vs multilingual approaches for each subtask, and found that the results of the comparison vary depending on the subtask. For subtask 1, multilingual transformer models demonstrate better average performance than monolingual models with translations. A similar effect was observed for subtask 2 and subtask 3 where multilingual settings achieved better performance than monolingual ones in all languages except English. We found the impact on bottleneck adapters to be unpredictable across tasks -- despite performing on average better for monolingual models in subtask~1, they were more beneficial for multilingual models in subtask~2 (and hindered monolingual performance). Finally, we presented post-competition findings, which suggest that subtask~2 would have benefited from a zero-shot prediction using multilingual models, while subtask~1 could have achieved much better results with the `translate-test' approach. Further analysis of this will be possible when test labels are released.

\section{Acknowledgments}
This work has been co-funded by the European Union under the Horizon Europe  vera.ai (grant 101070093) and Vigilant (grant 101073921) projects and the UK’s innovation agency (Innovate UK) grants 10039055 and 10039039.

\bibliography{anthology,custom}

\begin{thebibliography}{39}
\expandafter\ifx\csname natexlab\endcsname\relax\def\natexlab#1{#1}\fi

\bibitem[{Abu~Farha et~al.(2022)Abu~Farha, Oprea, Wilson, and
  Magdy}]{abu-farha-etal-2022-semeval}
Ibrahim Abu~Farha, Silviu~Vlad Oprea, Steven Wilson, and Walid Magdy. 2022.
\newblock \href {https://doi.org/10.18653/v1/2022.semeval-1.111}
  {{S}em{E}val-2022 task 6: i{S}arcasm{E}val, intended sarcasm detection in
  {E}nglish and {A}rabic}.
\newblock In \emph{Proceedings of the 16th International Workshop on Semantic
  Evaluation (SemEval-2022)}, pages 802--814, Seattle, United States.
  Association for Computational Linguistics.

\bibitem[{Adebara et~al.(2020)Adebara, Nagoudi, and
  Abdul~Mageed}]{adebara-etal-2020-translating}
Ife Adebara, El~Moatez~Billah Nagoudi, and Muhammad Abdul~Mageed. 2020.
\newblock \href {https://aclanthology.org/2020.wmt-1.42} {Translating similar
  languages: Role of mutual intelligibility in multilingual transformers}.
\newblock In \emph{Proceedings of the Fifth Conference on Machine Translation},
  pages 381--386, Online. Association for Computational Linguistics.

\bibitem[{Aghajanyan et~al.(2021)Aghajanyan, Gupta, Shrivastava, Chen,
  Zettlemoyer, and Gupta}]{aghajanyan-etal-2021-muppet}
Armen Aghajanyan, Anchit Gupta, Akshat Shrivastava, Xilun Chen, Luke
  Zettlemoyer, and Sonal Gupta. 2021.
\newblock \href {https://doi.org/10.18653/v1/2021.emnlp-main.468} {Muppet:
  Massive multi-task representations with pre-finetuning}.
\newblock In \emph{Proceedings of the 2021 Conference on Empirical Methods in
  Natural Language Processing}, pages 5799--5811, Online and Punta Cana,
  Dominican Republic. Association for Computational Linguistics.

\bibitem[{Aky{\"u}rek et~al.(2020)Aky{\"u}rek, Guo, Elanwar, Ishwar, Betke, and
  Wijaya}]{akyurek-etal-2020-multi}
Afra~Feyza Aky{\"u}rek, Lei Guo, Randa Elanwar, Prakash Ishwar, Margrit Betke,
  and Derry~Tanti Wijaya. 2020.
\newblock \href {https://doi.org/10.18653/v1/2020.acl-main.763} {Multi-label
  and multilingual news framing analysis}.
\newblock In \emph{Proceedings of the 58th Annual Meeting of the Association
  for Computational Linguistics}, pages 8614--8624, Online. Association for
  Computational Linguistics.

\bibitem[{Bapna and Firat(2019)}]{bapna-firat-2019-simple}
Ankur Bapna and Orhan Firat. 2019.
\newblock \href {https://doi.org/10.18653/v1/D19-1165} {Simple, scalable
  adaptation for neural machine translation}.
\newblock In \emph{Proceedings of the 2019 Conference on Empirical Methods in
  Natural Language Processing and the 9th International Joint Conference on
  Natural Language Processing (EMNLP-IJCNLP)}, pages 1538--1548, Hong Kong,
  China. Association for Computational Linguistics.

\bibitem[{Chalkidis et~al.(2021)Chalkidis, Fergadiotis, and
  Androutsopoulos}]{chalkidis-etal-2021-multieurlex}
Ilias Chalkidis, Manos Fergadiotis, and Ion Androutsopoulos. 2021.
\newblock \href {https://doi.org/10.18653/v1/2021.emnlp-main.559}
  {{M}ulti{EURLEX} - a multi-lingual and multi-label legal document
  classification dataset for zero-shot cross-lingual transfer}.
\newblock In \emph{Proceedings of the 2021 Conference on Empirical Methods in
  Natural Language Processing}, pages 6974--6996, Online and Punta Cana,
  Dominican Republic. Association for Computational Linguistics.

\bibitem[{Chan et~al.(2020)Chan, Schweter, and
  M{\"o}ller}]{chan-etal-2020-germans}
Branden Chan, Stefan Schweter, and Timo M{\"o}ller. 2020.
\newblock \href {https://doi.org/10.18653/v1/2020.coling-main.598}
  {{G}erman{'}s next language model}.
\newblock In \emph{Proceedings of the 28th International Conference on
  Computational Linguistics}, pages 6788--6796, Barcelona, Spain (Online).
  International Committee on Computational Linguistics.

\bibitem[{Conneau et~al.(2020)Conneau, Khandelwal, Goyal, Chaudhary, Wenzek,
  Guzm{\'a}n, Grave, Ott, Zettlemoyer, and
  Stoyanov}]{conneau-etal-2020-unsupervised}
Alexis Conneau, Kartikay Khandelwal, Naman Goyal, Vishrav Chaudhary, Guillaume
  Wenzek, Francisco Guzm{\'a}n, Edouard Grave, Myle Ott, Luke Zettlemoyer, and
  Veselin Stoyanov. 2020.
\newblock \href {https://doi.org/10.18653/v1/2020.acl-main.747} {Unsupervised
  cross-lingual representation learning at scale}.
\newblock In \emph{Proceedings of the 58th Annual Meeting of the Association
  for Computational Linguistics}, pages 8440--8451, Online. Association for
  Computational Linguistics.

\bibitem[{Da~San~Martino et~al.(2020)Da~San~Martino, Barr{\'o}n-Cede{\~n}o,
  Wachsmuth, Petrov, and Nakov}]{da-san-martino-etal-2020-semeval}
Giovanni Da~San~Martino, Alberto Barr{\'o}n-Cede{\~n}o, Henning Wachsmuth,
  Rostislav Petrov, and Preslav Nakov. 2020.
\newblock \href {https://doi.org/10.18653/v1/2020.semeval-1.186}
  {{S}em{E}val-2020 task 11: Detection of propaganda techniques in news
  articles}.
\newblock In \emph{Proceedings of the Fourteenth Workshop on Semantic
  Evaluation}, pages 1377--1414, Barcelona (online). International Committee
  for Computational Linguistics.

\bibitem[{Da~San~Martino et~al.(2019)Da~San~Martino, Yu, Barr{\'o}n-Cede{\~n}o,
  Petrov, and Nakov}]{da-san-martino-etal-2019-fine}
Giovanni Da~San~Martino, Seunghak Yu, Alberto Barr{\'o}n-Cede{\~n}o, Rostislav
  Petrov, and Preslav Nakov. 2019.
\newblock \href {https://doi.org/10.18653/v1/D19-1565} {Fine-grained analysis
  of propaganda in news article}.
\newblock In \emph{Proceedings of the 2019 Conference on Empirical Methods in
  Natural Language Processing and the 9th International Joint Conference on
  Natural Language Processing (EMNLP-IJCNLP)}, pages 5636--5646, Hong Kong,
  China. Association for Computational Linguistics.

\bibitem[{Devlin et~al.(2019)Devlin, Chang, Lee, and
  Toutanova}]{devlin-etal-2019-bert}
Jacob Devlin, Ming-Wei Chang, Kenton Lee, and Kristina Toutanova. 2019.
\newblock \href {https://doi.org/10.18653/v1/N19-1423} {{BERT}: Pre-training of
  deep bidirectional transformers for language understanding}.
\newblock In \emph{Proceedings of the 2019 Conference of the North {A}merican
  Chapter of the Association for Computational Linguistics: Human Language
  Technologies, Volume 1 (Long and Short Papers)}, pages 4171--4186,
  Minneapolis, Minnesota. Association for Computational Linguistics.

\bibitem[{Dimitrov et~al.(2021)Dimitrov, Bin~Ali, Shaar, Alam, Silvestri,
  Firooz, Nakov, and Da~San~Martino}]{dimitrov-etal-2021-semeval}
Dimitar Dimitrov, Bishr Bin~Ali, Shaden Shaar, Firoj Alam, Fabrizio Silvestri,
  Hamed Firooz, Preslav Nakov, and Giovanni Da~San~Martino. 2021.
\newblock \href {https://doi.org/10.18653/v1/2021.semeval-1.7}
  {{S}em{E}val-2021 task 6: Detection of persuasion techniques in texts and
  images}.
\newblock In \emph{Proceedings of the 15th International Workshop on Semantic
  Evaluation (SemEval-2021)}, pages 70--98, Online. Association for
  Computational Linguistics.

\bibitem[{Golbeck et~al.(2018)Golbeck, Mauriello, Auxier, Bhanushali, Bonk,
  Bouzaghrane, Buntain, Chanduka, Cheakalos, Everett, Falak, Gieringer, Graney,
  Hoffman, Huth, Ma, Jha, Khan, Kori, Lewis, Mirano, Mohn~IV, Mussenden,
  Nelson, Mcwillie, Pant, Shetye, Shrestha, Steinheimer, Subramanian, and
  Visnansky}]{golbeck18-fakenews}
Jennifer Golbeck, Matthew Mauriello, Brooke Auxier, Keval~H. Bhanushali,
  Christopher Bonk, Mohamed~Amine Bouzaghrane, Cody Buntain, Riya Chanduka,
  Paul Cheakalos, Jennine~B. Everett, Waleed Falak, Carl Gieringer, Jack
  Graney, Kelly~M. Hoffman, Lindsay Huth, Zhenya Ma, Mayanka Jha, Misbah Khan,
  Varsha Kori, Elo Lewis, George Mirano, William~T. Mohn~IV, Sean Mussenden,
  Tammie~M. Nelson, Sean Mcwillie, Akshat Pant, Priya Shetye, Rusha Shrestha,
  Alexandra Steinheimer, Aditya Subramanian, and Gina Visnansky. 2018.
\newblock \href {https://doi.org/10.1145/3201064.3201100} {Fake news vs satire:
  A dataset and analysis}.
\newblock In \emph{Proceedings of the 10th ACM Conference on Web Science},
  WebSci '18, page 17–21, New York, NY, USA. Association for Computing
  Machinery.

\bibitem[{Guibon et~al.(2019)Guibon, Ermakova, Seffih, Firsov, and
  Le~No{\'e}-Bienvenu}]{guibon2019multilingual}
Ga{\"e}l Guibon, Liana Ermakova, Hosni Seffih, Anton Firsov, and Guillaume
  Le~No{\'e}-Bienvenu. 2019.
\newblock \href {https://doi.org/10.1007/978-3-031-24340-0_29} {Multilingual
  fake news detection with satire}.
\newblock In \emph{{CICLing}: International Conference on Computational
  Linguistics and Intelligent Text Processing}.

\bibitem[{Gururangan et~al.(2020)Gururangan, Marasovi{\'c}, Swayamdipta, Lo,
  Beltagy, Downey, and Smith}]{gururangan-etal-2020-dont}
Suchin Gururangan, Ana Marasovi{\'c}, Swabha Swayamdipta, Kyle Lo, Iz~Beltagy,
  Doug Downey, and Noah~A. Smith. 2020.
\newblock \href {https://doi.org/10.18653/v1/2020.acl-main.740} {Don{'}t stop
  pretraining: Adapt language models to domains and tasks}.
\newblock In \emph{Proceedings of the 58th Annual Meeting of the Association
  for Computational Linguistics}, pages 8342--8360, Online. Association for
  Computational Linguistics.

\bibitem[{He et~al.(2020)He, Liu, Gao, and Chen}]{he2020deberta}
Pengcheng He, Xiaodong Liu, Jianfeng Gao, and Weizhu Chen. 2020.
\newblock \href {https://doi.org/10.48550/ARXIV.2006.03654} {{DeBERTa}:
  Decoding-enhanced {BERT} with disentangled attention}.
\newblock \emph{Computing Research Repository}, arXiv:2006.03654.
\newblock Version 6.

\bibitem[{He et~al.(2021)He, Liu, Ye, Tan, Ding, Cheng, Low, Bing, and
  Si}]{he-etal-2021-effectiveness}
Ruidan He, Linlin Liu, Hai Ye, Qingyu Tan, Bosheng Ding, Liying Cheng, Jiawei
  Low, Lidong Bing, and Luo Si. 2021.
\newblock \href {https://doi.org/10.18653/v1/2021.acl-long.172} {On the
  effectiveness of adapter-based tuning for pretrained language model
  adaptation}.
\newblock In \emph{Proceedings of the 59th Annual Meeting of the Association
  for Computational Linguistics and the 11th International Joint Conference on
  Natural Language Processing (Volume 1: Long Papers)}, pages 2208--2222,
  Online. Association for Computational Linguistics.

\bibitem[{Houlsby et~al.(2019)Houlsby, Giurgiu, Jastrzebski, Morrone,
  De~Laroussilhe, Gesmundo, Attariyan, and Gelly}]{houlsby2019parameter}
Neil Houlsby, Andrei Giurgiu, Stanislaw Jastrzebski, Bruna Morrone, Quentin
  De~Laroussilhe, Andrea Gesmundo, Mona Attariyan, and Sylvain Gelly. 2019.
\newblock \href {https://proceedings.mlr.press/v97/houlsby19a.html}
  {Parameter-efficient transfer learning for {NLP}}.
\newblock In \emph{Proceedings of the 36th International Conference on Machine
  Learning}, volume~97 of \emph{Proceedings of Machine Learning Research},
  pages 2790--2799. PMLR.

\bibitem[{Jurkiewicz et~al.(2020)Jurkiewicz, Borchmann, Kosmala, and
  Grali{\'n}ski}]{jurkiewicz-etal-2020-applicaai}
Dawid Jurkiewicz, {\L}ukasz Borchmann, Izabela Kosmala, and Filip
  Grali{\'n}ski. 2020.
\newblock \href {https://doi.org/10.18653/v1/2020.semeval-1.187} {{A}pplica{AI}
  at {S}em{E}val-2020 task 11: On {R}o{BERT}a-{CRF}, span {CLS} and whether
  self-training helps them}.
\newblock In \emph{Proceedings of the Fourteenth Workshop on Semantic
  Evaluation}, pages 1415--1424, Barcelona (online). International Committee
  for Computational Linguistics.

\bibitem[{Kaliyar et~al.(2021)Kaliyar, Goswami, and
  Narang}]{kaliyar2021fakebert}
Rohit~Kumar Kaliyar, Anurag Goswami, and Pratik Narang. 2021.
\newblock \href {https://doi.org/10.1007/s11042-020-10183-2} {Fake{BERT}: Fake
  news detection in social media with a {BERT}-based deep learning approach}.
\newblock \emph{Multimedia tools and applications}, 80(8):11765--11788.

\bibitem[{Kiesel et~al.(2019)Kiesel, Mestre, Shukla, Vincent, Adineh, Corney,
  Stein, and Potthast}]{kiesel-etal-2019-semeval}
Johannes Kiesel, Maria Mestre, Rishabh Shukla, Emmanuel Vincent, Payam Adineh,
  David Corney, Benno Stein, and Martin Potthast. 2019.
\newblock \href {https://doi.org/10.18653/v1/S19-2145} {{S}em{E}val-2019 task
  4: Hyperpartisan news detection}.
\newblock In \emph{Proceedings of the 13th International Workshop on Semantic
  Evaluation}, pages 829--839, Minneapolis, Minnesota, USA. Association for
  Computational Linguistics.

\bibitem[{Kim et~al.(2021)Kim, Shum, Susanj, and
  Hilgart}]{kim-etal-2021-revisiting}
Seungwon Kim, Alex Shum, Nathan Susanj, and Jonathan Hilgart. 2021.
\newblock \href {https://doi.org/10.18653/v1/2021.repl4nlp-1.11} {Revisiting
  pretraining with adapters}.
\newblock In \emph{Proceedings of the 6th Workshop on Representation Learning
  for NLP (RepL4NLP-2021)}, pages 90--99, Online. Association for Computational
  Linguistics.

\bibitem[{Kuratov and Arkhipov(2019)}]{kuratov-19-rubert}
Yuri Kuratov and Mikhail Arkhipov. 2019.
\newblock \href {https://doi.org/10.48550/ARXIV.1905.07213} {Adaptation of deep
  bidirectional multilingual transformers for russian language}.
\newblock \emph{Computing Research Repository}, arXiv:1905.07213.
\newblock Version 1.

\bibitem[{Lan et~al.(2019)Lan, Chen, Goodman, Gimpel, Sharma, and
  Soricut}]{lan2019albert}
Zhenzhong Lan, Mingda Chen, Sebastian Goodman, Kevin Gimpel, Piyush Sharma, and
  Radu Soricut. 2019.
\newblock \href {https://doi.org/10.48550/ARXIV.1909.11942} {{ALBERT}: A lite
  {BERT} for self-supervised learning of language representations}.
\newblock \emph{Computing Research Repository}, arXiv:1909:11942.
\newblock Version 6.

\bibitem[{Liu and Xie(2021)}]{liu2021research}
Hongying Liu and Ling Xie. 2021.
\newblock \href {https://doi.org/10.1109/ICESIT53460.2021.9696851} {Research on
  sarcasm detection of news headlines based on {Bert-LSTM}}.
\newblock In \emph{2021 IEEE international conference on emergency science and
  information technology (ICESIT)}, pages 89--92. IEEE.

\bibitem[{Liu et~al.(2019{\natexlab{a}})Liu, Guo, Mays, Betke, and
  Wijaya}]{liu-etal-2019-detecting}
Siyi Liu, Lei Guo, Kate Mays, Margrit Betke, and Derry~Tanti Wijaya.
  2019{\natexlab{a}}.
\newblock \href {https://doi.org/10.18653/v1/K19-1047} {Detecting frames in
  news headlines and its application to analyzing news framing trends
  surrounding {U}.{S}. gun violence}.
\newblock In \emph{Proceedings of the 23rd Conference on Computational Natural
  Language Learning (CoNLL)}, pages 504--514, Hong Kong, China. Association for
  Computational Linguistics.

\bibitem[{Liu et~al.(2019{\natexlab{b}})Liu, Ott, Goyal, Du, Joshi, Chen, Levy,
  Lewis, Zettlemoyer, and Stoyanov}]{liu2019roberta}
Yinhan Liu, Myle Ott, Naman Goyal, Jingfei Du, Mandar Joshi, Danqi Chen, Omer
  Levy, Mike Lewis, Luke Zettlemoyer, and Veselin Stoyanov. 2019{\natexlab{b}}.
\newblock \href {https://doi.org/10.48550/ARXIV.1907.11692} {{RoBERTa}: A
  robustly optimized bert pretraining approach}.
\newblock \emph{Computing Research Repository}, arXiv:1907.11692.
\newblock Version 1.

\bibitem[{Martin et~al.(2020)Martin, Muller, Ortiz~Su{\'a}rez, Dupont, Romary,
  de~la Clergerie, Seddah, and Sagot}]{martin-etal-2020-camembert}
Louis Martin, Benjamin Muller, Pedro~Javier Ortiz~Su{\'a}rez, Yoann Dupont,
  Laurent Romary, {\'E}ric de~la Clergerie, Djam{\'e} Seddah, and Beno{\^\i}t
  Sagot. 2020.
\newblock \href {https://doi.org/10.18653/v1/2020.acl-main.645} {{C}amem{BERT}:
  a tasty {F}rench language model}.
\newblock In \emph{Proceedings of the 58th Annual Meeting of the Association
  for Computational Linguistics}, pages 7203--7219, Online. Association for
  Computational Linguistics.

\bibitem[{McHardy et~al.(2019)McHardy, Adel, and
  Klinger}]{mchardy-etal-2019-adversarial}
Robert McHardy, Heike Adel, and Roman Klinger. 2019.
\newblock \href {https://doi.org/10.18653/v1/N19-1069} {Adversarial training
  for satire detection: Controlling for confounding variables}.
\newblock In \emph{Proceedings of the 2019 Conference of the North {A}merican
  Chapter of the Association for Computational Linguistics: Human Language
  Technologies, Volume 1 (Long and Short Papers)}, pages 660--665, Minneapolis,
  Minnesota. Association for Computational Linguistics.

\bibitem[{Pandey and Singh(2022)}]{pandey2022bert}
Rajnish Pandey and Jyoti~Prakash Singh. 2022.
\newblock \href {https://doi.org/10.1007/s10844-022-00755-z} {{BERT-LSTM} model
  for sarcasm detection in code-mixed social media post}.
\newblock \emph{Journal of Intelligent Information Systems}, pages 1--20.

\bibitem[{Pfeiffer et~al.(2020{\natexlab{a}})Pfeiffer, R{\"u}ckl{\'e}, Poth,
  Kamath, Vuli{\'c}, Ruder, Cho, and Gurevych}]{pfeiffer-etal-2020-adapterhub}
Jonas Pfeiffer, Andreas R{\"u}ckl{\'e}, Clifton Poth, Aishwarya Kamath, Ivan
  Vuli{\'c}, Sebastian Ruder, Kyunghyun Cho, and Iryna Gurevych.
  2020{\natexlab{a}}.
\newblock \href {https://doi.org/10.18653/v1/2020.emnlp-demos.7}
  {{A}dapter{H}ub: A framework for adapting transformers}.
\newblock In \emph{Proceedings of the 2020 Conference on Empirical Methods in
  Natural Language Processing: System Demonstrations}, pages 46--54, Online.
  Association for Computational Linguistics.

\bibitem[{Pfeiffer et~al.(2020{\natexlab{b}})Pfeiffer, Vuli{\'c}, Gurevych, and
  Ruder}]{pfeiffer-etal-2020-mad}
Jonas Pfeiffer, Ivan Vuli{\'c}, Iryna Gurevych, and Sebastian Ruder.
  2020{\natexlab{b}}.
\newblock \href {https://doi.org/10.18653/v1/2020.emnlp-main.617} {{MAD-X}:
  {A}n {A}dapter-{B}ased {F}ramework for {M}ulti-{T}ask {C}ross-{L}ingual
  {T}ransfer}.
\newblock In \emph{Proceedings of the 2020 Conference on Empirical Methods in
  Natural Language Processing (EMNLP)}, pages 7654--7673, Online. Association
  for Computational Linguistics.

\bibitem[{Piskorski et~al.(2023)Piskorski, Stefanovitch, Da~San~Martino, and
  Nakov}]{semeval2023task3}
Jakub Piskorski, Nicolas Stefanovitch, Giovanni Da~San~Martino, and Preslav
  Nakov. 2023.
\newblock Semeval-2023 task 3: Detecting the category, the framing, and the
  persuasion techniques in online news in a multi-lingual setup.
\newblock In \emph{Proceedings of the 17th International Workshop on Semantic
  Evaluation}, SemEval 2023, Toronto, Canada.

\bibitem[{Tian et~al.(2021)Tian, Gui, Li, Yan, and Xiao}]{tian-etal-2021-mind}
Junfeng Tian, Min Gui, Chenliang Li, Ming Yan, and Wenming Xiao. 2021.
\newblock \href {https://doi.org/10.18653/v1/2021.semeval-1.150} {{M}in{D} at
  {S}em{E}val-2021 task 6: Propaganda detection using transfer learning and
  multimodal fusion}.
\newblock In \emph{Proceedings of the 15th International Workshop on Semantic
  Evaluation (SemEval-2021)}, pages 1082--1087, Online. Association for
  Computational Linguistics.

\bibitem[{Tsur et~al.(2015)Tsur, Calacci, and Lazer}]{tsur-etal-2015-frame}
Oren Tsur, Dan Calacci, and David Lazer. 2015.
\newblock \href {https://doi.org/10.3115/v1/P15-1157} {A frame of mind: Using
  statistical models for detection of framing and agenda setting campaigns}.
\newblock In \emph{Proceedings of the 53rd Annual Meeting of the Association
  for Computational Linguistics and the 7th International Joint Conference on
  Natural Language Processing (Volume 1: Long Papers)}, pages 1629--1638,
  Beijing, China. Association for Computational Linguistics.

\bibitem[{Wang and Banko(2021)}]{wang-banko-2021-practical}
Cindy Wang and Michele Banko. 2021.
\newblock \href {https://doi.org/10.18653/v1/2021.naacl-industry.16} {Practical
  transformer-based multilingual text classification}.
\newblock In \emph{Proceedings of the 2021 Conference of the North American
  Chapter of the Association for Computational Linguistics: Human Language
  Technologies: Industry Papers}, pages 121--129, Online. Association for
  Computational Linguistics.

\bibitem[{Wu and Dredze(2020)}]{wu-dredze-2020-languages}
Shijie Wu and Mark Dredze. 2020.
\newblock \href {https://doi.org/10.18653/v1/2020.repl4nlp-1.16} {Are all
  languages created equal in multilingual {BERT}?}
\newblock In \emph{Proceedings of the 5th Workshop on Representation Learning
  for NLP}, pages 120--130, Online. Association for Computational Linguistics.

\bibitem[{Xenouleas et~al.(2022)Xenouleas, Tsoukara, Panagiotakis, Chalkidis,
  and Androutsopoulos}]{xenouleas2022realistic}
Stratos Xenouleas, Alexia Tsoukara, Giannis Panagiotakis, Ilias Chalkidis, and
  Ion Androutsopoulos. 2022.
\newblock \href {https://doi.org/10.1145/3549737.3549760} {Realistic zero-shot
  cross-lingual transfer in legal topic classification}.
\newblock In \emph{Proceedings of the 12th Hellenic Conference on Artificial
  Intelligence}, SETN '22, New York, NY, USA. Association for Computing
  Machinery.

\bibitem[{Yang et~al.(2019)Yang, Dai, Yang, Carbonell, Salakhutdinov, and
  Le}]{yang2019xlnet}
Zhilin Yang, Zihang Dai, Yiming Yang, Jaime~G. Carbonell, Ruslan Salakhutdinov,
  and Quoc~V. Le. 2019.
\newblock \href
  {https://proceedings.neurips.cc/paper/2019/file/dc6a7e655d7e5840e66733e9ee67cc69-Paper.pdf}
  {Xlnet: Generalized autoregressive pretraining for language understanding}.
\newblock In \emph{Neural Information Processing Systems}.

\end{thebibliography}
\bibliographystyle{acl_natbib}

\appendix
\section{Language Models Used}

The models used in each subtask are shown in Table \ref{tab:model-list}.


\begin{table*}[]
\centering
\scalebox{0.8}{
    \begin{tabular}{l l l p{0.2cm}p{0.2cm}p{0.2cm}}
    \toprule
    Language     & Huggingface Model Name             & Publication                & \multicolumn{3}{l}{Subtasks} \\ \midrule
    English      & \texttt{bert-base-cased}                    & \citet{devlin-etal-2019-bert}      & 1 &&         \\ 
    English & \texttt{RoBERTa-base}                            & \citet{liu2019roberta}              & && 3$^\star$             \\
    English & \texttt{RoBERTa-large}                            & \citet{liu2019roberta}              & & 2 &              \\
    English & \texttt{MUPPET-large}                            & \citet{aghajanyan-etal-2021-muppet}   & & 2$^\star$ &              \\
    
    French       & \texttt{camembert-base}                     & \citet{martin-etal-2020-camembert} & 1 &&             \\
    German       & \texttt{deepset/gbert-base}                 & \citet{chan-etal-2020-germans}     & 1   &&             \\
    Italian      & \texttt{dbmdz/bert-base-italian-cased}      & -                          & 1 &&               \\
    Polish       & \texttt{dkleczek/bert-base-polish-cased-v1} & -                          & 1  &&          \\
    Russian      & \texttt{DeepPavlov/rubert-base-cased}       & \citet{kuratov-19-rubert}          & 1 &&                \\
    Multilingual & \texttt{bert-base-multilingual-cased}       & \citet{devlin-etal-2019-bert}      & 1$^\star$&&3$^\star$                \\
    Multilingual & \texttt{xlm-RoBERTa-base} & \citet{conneau-etal-2020-unsupervised} & 1 &  & 3 \\
    Multilingual & \texttt{xlm-RoBERTa-large} & \citet{conneau-etal-2020-unsupervised} & & 2$^\star$ & \\
    \bottomrule
    \end{tabular}
}
\caption{Models used in each subtask. $^\star$ denotes the models used in the final submission. Based on model selection of \citet{chalkidis-etal-2021-multieurlex}.}
\label{tab:model-list}
\end{table*}



\section{Article Preprocessing}\label{ap:1}
Articles were preprocessed with the following steps, for all languages:
\begin{itemize}
    \item add a full stop at the end of the title.
    \item remove duplicate sentences directly following each other;
    \item remove the @ symbol from twitter handles;
    \item remove hyperlinks to websites and images;
\end{itemize}
English articles were further preprocessed:
\begin{itemize}
    \item remove lines indicating the possibility to share the article on different social media platforms;
    \item remove sentences suggesting the user to take part in  online polls, comments, or advertisements;
    \item remove sentences indicating the terms of use;
    \item remove sentences indicating the licenses and containing phrases such as `reprinted with permission', `posted with permission' and `all rights reserved';
    \item remove sentences relating to the article author biographies
\end{itemize}

\section{Subtask~1}
\subsection{Language-specific performance after each epoch on development set}\label{ap:4}
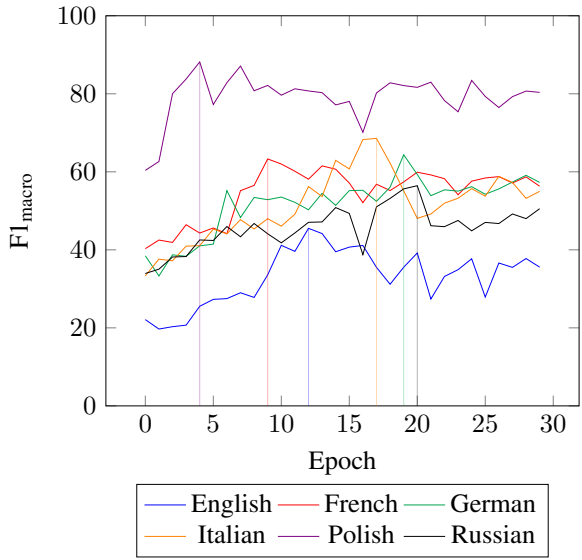
\begin{figure}[!h]
\resizebox{\columnwidth}{!}{%
    \begin{tikzpicture}
        \begin{axis}[
            no markers,
            xlabel=Epoch,
            ylabel={\fmacro},
            ymin=0,
            ymax=100,
            legend style={at={(0.5,-0.2)},anchor=north},
            legend columns=3,
            cycle list={
                blue,
                red,
                Green,
                orange,
                violet,
                black
            }
        ]
        \addplot file {figures/st1_langs/English.txt}; \addlegendentry{English}
        \addplot file {figures/st1_langs/French.txt}; \addlegendentry{French}
        \addplot file {figures/st1_langs/German.txt}; \addlegendentry{German}
        \addplot file {figures/st1_langs/Italian.txt}; \addlegendentry{Italian}
        \addplot file {figures/st1_langs/Polish.txt}; \addlegendentry{Polish}
        \addplot file {figures/st1_langs/Russian.txt}; \addlegendentry{Russian}

        \addplot[blue, opacity=0.3] coordinates {(12,0) (12,45.51)};
        \addplot[red, opacity=0.3] coordinates {(9,0) (9,63.30)};
        \addplot[Green, opacity=0.3] coordinates {(19,0) (19,64.37)};
        \addplot[orange, opacity=0.3] coordinates {(17,0) (17,68.56)};
        \addplot[violet, opacity=0.3] coordinates {(4,0) (4,88.16)};
        \addplot[black, opacity=0.3] coordinates {(20,0) (20,56.44)};
        \end{axis}
    \end{tikzpicture}
    }
    \caption{Validation \fmacro of each language over time, with maximal epoch indicated.}
    \label{fig1:st1_train}
\end{figure}

Figure~\ref{fig1:st1_train} shows \fmacro scores on the held-out development set for one of the finetuned transformer models. As can be seen, Polish reaches its best performance quite early on, while German and Russian need more than 17 epochs to achieve the best score.

\section{Subtask 2}\label{ap:2}
Hyperparameters for Subtask 2 TAPT are shown in Table \ref{tab:st2_tapt_hyperparams}.
The detailed cross-validation results are shown in Table \ref{tab:st2_full_crossvalidation_results}.

\begin{table}[]
    \centering
    \scalebox{0.8}{
    \begin{tabular}{l c}
       \toprule
       Epochs & 60 \\
       Effective Batch Size & 128 \\ 
       Max learning rate & 1e-4 \\
       Warmup ratio & 0.06, linear \\
       Learning rate decay & linear \\
       Optimiser & AdamW \\ 
       Adam epsilon & 1e-6 \\
       Adam beta weights & 0.9, 0.98 \\
       Weight decay & 0.01  \\
       \bottomrule
    \end{tabular}
    }
    \vspace{-2mm}
    \caption{Subtask 2: TAPT Hyperparameters}
    \vspace{-4mm}
    \label{tab:st2_tapt_hyperparams}
\end{table}

\subsection{Post-competition Findings}
Table \ref{tab:st2_post} shows the full set of post-competition results. The multilingual model outperforms our official monolingual submission for test set English, Greek, and Spanish, but is much worse for Georgian. Conversely, our monolingual model does better than our official multilingual submission for French and Italian. This suggests that neither monolingual nor multilingual models are consistently better than the other across all languages. 

Traditional zero-shot cross-lingual experiments focus on finetuning a multilingual model on a single source language and testing in a target language. In our situation, we jointly finetune on multiple languages, which encourages the model to retain multilingual representations, even for languages not in its training, thus improving its zero-shot cross-lingual capabilities. This may help to explain the improved performance of our multilingual model. However, it is important to note that these results are not representative of true zero-shot classification, since our multilingual model did perform task-adaptive pre-training on articles from the surprise languages. Unfortunately, because the organisers have not released labels for the test set, we are unable to perform error analysis. As mentioned in the main section, the small size of the test set also makes it difficult to draw firm conclusions on whether translate-test is better than multilingual zero-shot classification.  

\begin{table}[]
    \centering
    \scalebox{0.8}{
        \begin{tabular}{l l l}
       \toprule
       Language & Multilingual & Monolingual \\ 
       \midrule
       English  & \textbf{58.475} & 57.895 (1) \\
       French & 53.425 (3) & \textbf{54.181} \\
       German & \textbf{65.251} (3) & 62.069\\
       Italian & 57.079 (7) & \textbf{60.577}\\
       Polish & \textbf{64.516} (2) & 63.581\\
       Russian & \textbf{44.144} (2) & 40.800\\
       \midrule 
       Spanish & \textbf{52.023} & 50.829 (3)\\
       Greek & \textbf{58.000} & 54.630 (1)\\
       Georgian & 60.870 & \textbf{65.421} (1)\\
       \bottomrule
       \end{tabular}
    }
    \vspace{-2mm}
    \caption{Subtask 2: \fmicro Perfomance on test set - post-competition comparison. () indicates ranking for official submissions.}
    \vspace{-4mm}
    \label{tab:st2_post}
    
\end{table}

\begin{table*}
\centering
\scalebox{0.8}{
    \begin{tabular}{l c c c c c c c}
    \toprule
    \textbf{Monolingual English} & EN & DE & FR & IT & PO & RU & Overall \fmicro \\

    \midrule

    RoBERTa-Large & 68.4 ± 2.0 & 63.5 ± 2.0 & 57.9 ± 2.9 & 60.9 ± 0.2 & 65.8 ± 3.4 & 54.5 ± 2.7 & 63.6 ± 0.1\\
    MUPPET-Large & 70.4 ± 2.0 & 62.1 ± 3.7 & 59.0 ± 0.9 & 58.3 ± 1.5 & 65.7 ± 0.9 & 52.9 ± 1.7 & 63.5 ± 0.7\\
    MUPPET-Large + Adapters & 68.0 ± 1.0 & 59.5 ± 1.6 & 54.5 ± 1.9 & 58.0 ± 0.7 & 61.9 ± 2.1 & 51.0 ± 3.7 & 61.1 ± 0.9 \\
    \midrule
    \textbf{Multilingual Models} & & & & & & & \\
    \midrule
    XLM-R & 68.3 ± 1.4 & 64.4 ± 1.4 & 58.5 ± 0.7 & 60.6 ± 0.5 & 66.5 ± 3.3 & 54.9 ± 2.0 & 64.0 ± 1.2\\
    XLM-R + Adapters & 69.0 ± 1.2 & 64.0 ± 1.4 & 58.4 ± 3.0 & 61.3 ± 0.9 & 67.4 ± 1.0 & 53.1 ± 2.7 & 64.3 ± 0.4\\
    XLM-R + TAPT + Adapters & 68.2 ± 0.9 & 65.0 ± 1.8 & 58.5 ± 2.8 & 61.0 ± 0.6 & 66.7 ± 3.0 & 55.7 ± 3.1 & 64.2 ± 0.3\\

    \midrule

    XLM-R (no class weights) & 68.8 ± 1.7 & 57.1 ± 2.2 & 61.6 ± 4.0 & 61.7 ± 1.2 & 67.4 ± 2.1 & 57.0 ± 2.2 & 65.1 ± 0.2\\
    \bottomrule
    
    \end{tabular}
}

\caption{Full version of Subtask 2 cross-validation results. Comparison of averaged \fmicro scores on 3-fold cross-validation (merged training and organiser-dev set). All models have class-weighting, except where indicated otherwise.}
\vspace{-4mm}

\label{tab:st2_full_crossvalidation_results}
\end{table*}

\section{Subtask 3}\label{ap:3}
\subsection{Training With vs. Without Non-Labelled Examples} \label{ap:classless}
In the combined training set across all the languages, there are 9,280 paragraphs that do not have a label. Although it is expected that this also occurs on the test set, the organizer's baseline approach discards these train samples, so it never explicitly trains on unlabelled examples.
Table \ref{tab:st3-labelled-vs-unlabelled-size} displays the sizes of the train set for each language without adding non-labelled examples vs. adding them. Table \ref{tab:st3-labelled-vs-unlabelled-f1micro} shows the \fmicro results of both approaches, with means and stds computed over three random seed initialisations. Note that adding the non-labelled examples contributes to a considerable increase in performance for all languages, particularly English, which is also the language that had the biggest increase in train set size.

\begin{table}
\centering
\scalebox{0.8}{
    \begin{tabular}{l r r}
    \toprule
    \multirow{2}{*}{Language} & \multicolumn{2}{c}{Training Set Size}                                                  \\ \cmidrule(lr){2-3}
    & \multicolumn{1}{l}{Without Non-Labelled} & \multicolumn{1}{l}{With Non-Labelled} \\ \midrule
    English & 3760                                  & 9498 (+152\%)  \\
    French  & 1693                                  & 2259 (+33\%)                       \\
    German  & 1252                                  & 1555 (+24\%)                       \\
    Italian & 1745                                  & 2623 (+50\%)                       \\
    Polish  & 1232                                  & 2310 (+32\%)                       \\
    Russian & 1245                                  & 1962 (+57\%)                       \\ \bottomrule
    \end{tabular}
}
\caption{Subtask 3 train set sizes for each language without and with the addition of examples that weren't assigned a class during labelling.}
\label{tab:st3-labelled-vs-unlabelled-size}
\end{table}

\begin{table}
\centering
\scalebox{0.8}{
\begin{tabular}{@{}lcc@{}}
\toprule
\multirow{2}{*}{Language} & \multicolumn{2}{c}{\fmicro $\pm$ 1 std}                                                                       \\ \cmidrule(lr){2-3}
 & Without Classless                            & With Classless       \\ \midrule
English                      & 27.1  ±  1.0                                 & \textbf{36.2 ± 0.3}  \\
French                       & 41.3  ± 0.1                                  & \textbf{43.4 ± 0.4}  \\
German                       & 40.8  ± 0.1                         & \textbf{40.9 ± 0.8}  \\
Italian                      & 44.1  ± 0.6                                  & \textbf{47.5 ± 0.4}  \\
Polish                       & 27.8  ± 0.9                                  & \textbf{30.2 ± 0.1}  \\
Russian                      & 35.7  ±  0.9                        & \textbf{37.5  ± 2.0} \\ \bottomrule
\end{tabular}
}
\caption{Subtask 3 \fmicro for best model configurations for each language with and without the addition of classless examples. Best \fmicro per language are marked as \textbf{bold}.}
\label{tab:st3-labelled-vs-unlabelled-f1micro}
\vspace{-4mm}
\end{table}

\subsection{Development Set Fine-grained Results}
Table \ref{tab:st3-error-analysis} shows the fine-grained results for the English official development set. Results are obtained from the best random seed over three runs. Although \textit{Appeal\_to\_Time}, \textit{Appeal\_to\_Values}, \textit{Consequential\_Oversimplification} and \textit{Questioning\_the\_Reputation} classes do not have a single example in development set, there are six other classes in which we also obtain $0.0$ F1-Score, namely \textit{Appeal\_to\_Hypocrisy}, \textit{Appeal\_to\_Popularity}, \textit{Obfuscation-Vagueness-Confusion}, \textit{Red\_Herring}, \textit{Straw\_Man}
and \textit{Whataboutism}, although together they account for only 5\% of the development set. The three biggest classes, \textit{Loaded\_Language}, \textit{Name\_Calling-Labeling} and \textit{Doubt} account for 29\%, 15\% and 11\% of the development set, respectively, thus having a large impact on \fmicro. 

\begin{table}
\centering
\scalebox{0.8}{
    \begin{tabular}{lrcrc}
    \toprule
    \multicolumn{5}{c}{Final Submission}                                                                                                   \\
             & \multicolumn{1}{l}{Test \fmicro} & \multicolumn{1}{l}{Place} & \multicolumn{1}{l}{Test \fmacro} & \multicolumn{1}{l}{Place} \\ \midrule
    English  & 36.802                            & 2                        & 17.194                            & 2                        \\
    French   & 41.436                            & 4                        & 32.424                            & 1                        \\
    German   & 44.726                            & 6                        & 23.679                            & 3                        \\
    Italian  & 52.494                            & 3                        & 28.22                             & 1                        \\
    Polish   & 34.7                              & 7                        & 19.102                            & 4                        \\
    Russian  & 31.841                            & 5                        & 20.522                            & 2                        \\ \hdashline
    Greek    & 17.426                            & 7                        & 11.028                            & 8                        \\
    Spanish  & 27.497                            & 9                        & 13.042                            & 8                        \\
    Georgian & 24.911                            & 10                       & 29.553                            & 4                        \\ \bottomrule
    \end{tabular}
}
\caption{Subtask 3 final submission \fmicro and \fmacro and our placement according to both of them.}
\label{tab:st3-final-results-micromacro}
\vspace{-4mm}
\end{table}

\begin{table*}
\centering
\scalebox{0.8}{
\begin{tabular}{@{}lrrrr@{}}
\toprule
Class                             & \multicolumn{1}{l}{Precision} & \multicolumn{1}{l}{Recall} & \multicolumn{1}{l}{F1-Score} & \multicolumn{1}{l}{Samples} \\ \midrule
Appeal\_to\_Authority             & 0.11                          & 0.07                       & 0.09                         & 28                          \\
Appeal\_to\_Fear-Prejudice        & 0.39                          & 0.23                       & 0.29                         & 137                         \\
Appeal\_to\_Hypocrisy             & 0                             & 0                          & 0                            & 8                           \\
Appeal\_to\_Popularity            & 0                             & 0                          & 0                            & 34                          \\
Appeal\_to\_Time                  & 0                             & 0                          & 0                            & 0                           \\
Appeal\_to\_Values                & 0                             & 0                          & 0                            & 0                           \\
Causal\_Oversimplification        & 0.03                          & 0.04                       & 0.04                         & 24                          \\
Consequential\_Oversimplification & 0                             & 0                          & 0                            & 0                           \\
Conversation\_Killer              & 0.11                          & 0.28                       & 0.16                         & 25                          \\
Doubt                             & 0.26                          & 0.36                       & 0.3                          & 187                         \\
Exaggeration-Minimisation         & 0.21                          & 0.34                       & 0.26                         & 115                         \\
False\_Dilemma-No\_Choice         & 0.26                          & 0.16                       & 0.2                          & 63                          \\
Flag\_Waving                      & 0.34                          & 0.49                       & 0.4                          & 96                          \\
Guilt\_by\_Association            & 0.33                          & 0.25                       & 0.29                         & 4                           \\
Loaded\_Language                  & 0.39                          & 0.64                       & 0.48                         & 483                         \\
Name\_Calling-Labeling            & 0.42                          & 0.69                       & 0.52                         & 250                         \\
Obfuscation-Vagueness-Confusion   & 0                             & 0                          & 0                            & 13                          \\
Questioning\_the\_Reputation      & 0                             & 0                          & 0                            & 0                           \\
Red\_Herring                      & 0                             & 0                          & 0                            & 19                          \\
Repetition                        & 0.12                          & 0.24                       & 0.16                         & 141                         \\
Slogans                           & 0.21                          & 0.43                       & 0.29                         & 28                          \\
Straw\_Man                        & 0                             & 0                          & 0                            & 9                           \\
Whataboutism                      & 0                             & 0                          & 0                            & 2                           \\ \midrule
micro avg                         & 0.31                          & 0.44                       & 0.36                         & 1666                        \\
macro avg                         & 0.14                          & 0.18                       & 0.15                         & 1666                        \\ \bottomrule
\end{tabular}
}
\caption{Subtask~3 fine-grained results for the English development set.}
    \label{tab:st3-error-analysis}
    
\end{table*}

\subsection{Full Leaderboard Results} \label{ap:s3-full-leaderboard}
Table \ref{tab:st3-final-results-micromacro} shows our full final submission scores and placements according to both \fmicro and \fmacro. As we previously point out in section \ref{sec:st3-system-overview}, we aimed towards a model capable of identifying all the 23 classes, thus having high \fmacro, even though the main metric for the subtask is \fmicro. We believe that a realistic application of a model for this particular label scheme should not disregard under-represented classes, otherwise they should simply be removed from the label scheme. Although our placings according to \fmacro are considerably higher, we acknowledge that because the main metric for the subtask is not \fmacro, other teams' submissions are likely not focusing on maximizing it, thus making their scores lower on average.

\end{document}